\documentclass[10pt,twocolumn,letterpaper]{article}

\usepackage{wacv}
\usepackage{times}
\usepackage{epsfig}
\usepackage{graphicx}
\usepackage{amsmath}
\usepackage{amssymb}

\usepackage{amssymb}

\usepackage{adjustbox}
\usepackage{rotating}

\usepackage{xcolor}

\wacvfinalcopy 

\ifwacvfinal
\def\assignedStartPage{1}
\fi

\ifwacvfinal
\usepackage[breaklinks=true,bookmarks=false]{hyperref}
\else
\usepackage[pagebackref=true,breaklinks=true,colorlinks,bookmarks=false]{hyperref}
\fi

\ifwacvfinal
\else
\fi

\newcommand{\datasetname}{DRIV100}
\newcommand{\datasetnamelong}{Diverse Roadscenes from Internet Videos 100}

\makeatletter
\g@addto@macro{\UrlBreaks}{\UrlOrds}
\makeatother

\begin{document}

\title{\datasetname: In-The-Wild Multi-Domain Dataset and Evaluation\\ for Real-World Domain Adaptation of Semantic Segmentation}

\author{Haruya Sakashita\thanks{The first two authors contribute equally to this work.}\\\
Osaka University\\
{\tt\small sakashita.haruya@ist.osaka-u.ac.jp}
\and
Christoph Flothow\footnotemark[1]\\
Technische Universität Darmstadt\\
{\tt\small christoph.flothow@stud.tu-darmstadt.de}
\and
Noriko Takemura\\
Osaka University\\
{\tt\small takemura@ids.osaka-u.ac.jp}
\and
Yusuke Sugano\\
The University of Tokyo\\
{\tt\small sugano@iis.u-tokyo.ac.jp}
}

\maketitle

\begin{abstract}
Together with the recent advances in semantic segmentation, many domain adaptation methods have been proposed to overcome the domain gap between training and deployment environments.
However, most previous studies use limited combinations of source/target datasets, and domain adaptation techniques have never been thoroughly evaluated in a more challenging and diverse set of target domains.
This work presents a new multi-domain dataset \datasetname~for benchmarking domain adaptation techniques on in-the-wild road-scene videos collected from the Internet.
The dataset consists of pixel-level annotations for 100 videos selected to cover diverse scenes/domains based on two criteria; human subjective judgment and an anomaly score judged using an existing road-scene dataset.
We provide multiple manually labeled ground-truth frames for each video, enabling a thorough evaluation of video-level domain adaptation where each video independently serves as the target domain.
Using the dataset, we quantify domain adaptation performances of state-of-the-art methods and clarify the potential and novel challenges of domain adaptation techniques.
The dataset is available at \url{https://doi.org/10.5281/zenodo.4389243}.
\end{abstract}

\section{Introduction}

Semantic segmentation has been one of the most active research topics in computer vision over the past years and is considered essential in many application scenarios, including autonomous driving.
Large-scale segmentation datasets from real-world image datasets~\cite{camvid,kitti,cityscapes,BDD100K,MVD2017} to fully synthetic ones~\cite{HernandezBMVC17,GTA5} enabled the advancement in deep learning-based semantic segmentation models~\cite{FCN,segnet,PSP,U-Net,chen2017deeplab}, and state-of-the-art models show a remarkable performance on these datasets.
However, it is still quite challenging to fully avoid the issue of domain biases between training and testing environments.
To overcome the domain gap and fully utilize synthetic training data, many recent works have focused on the unsupervised domain adaptation of semantic segmentation~\cite{sankaranarayanan2018learning,luo2019taking,zou2019confidence}.
The goal is to train a segmentation model on the target domain by using only annotated training samples on the source domain.
Despite its progress, however, most state-of-the-art domain adaptation methods used only limited combinations of source/target datasets for evaluation.

\begin{figure}[t]
  \centering
  \includegraphics[width=1.0\linewidth]{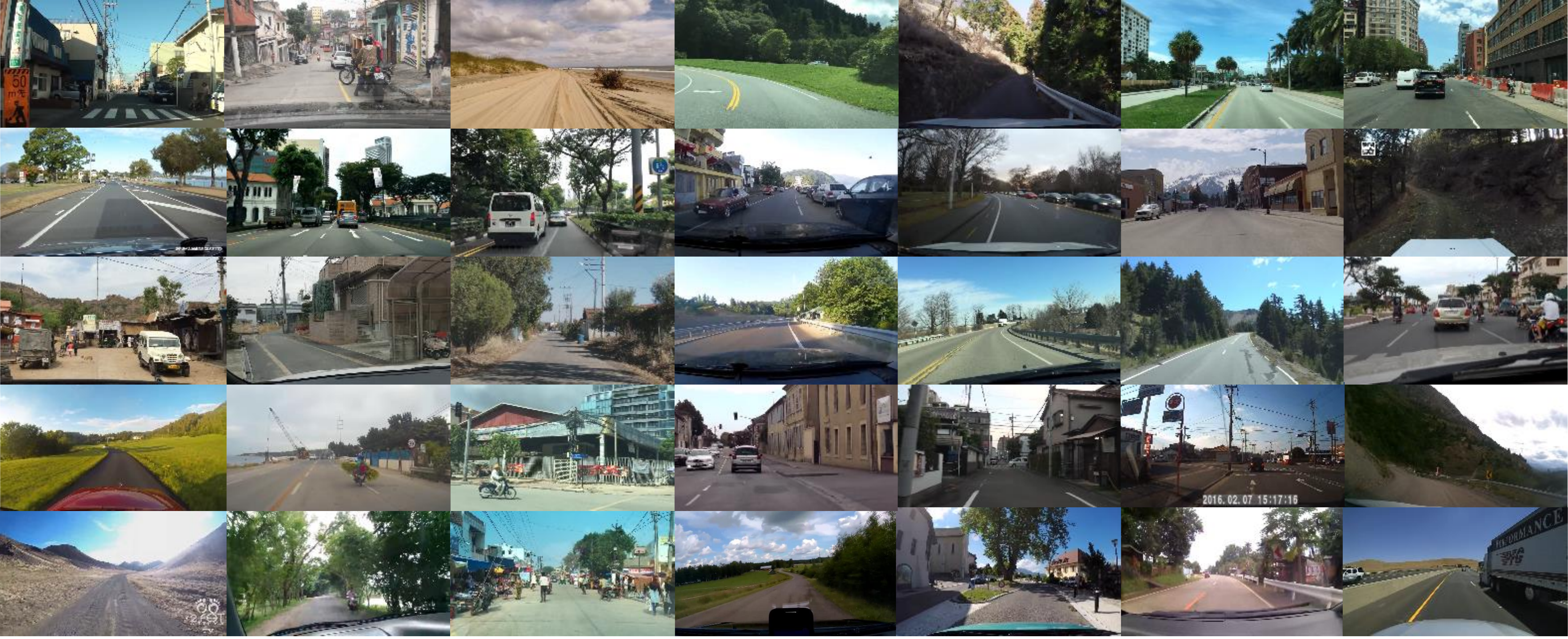} 
  \caption{Examples of adaptation target videos in our \datasetname~dataset}
  \label{fig:DRIV-100thumb}
\end{figure}

On the other hand, there have been some recent attempts to create semantic segmentation datasets in more diverse scenes~\cite{CrossCity,zendel2018wilddash,BDD100K,MVD2017}.
Although such datasets serve as a valuable resource for evaluating domain adaptation methods, there are still some critical limitations as the target for domain adaptation.
Firstly, the definition of domain is not clear in these datasets, and they implicitly assume a dataset-level domain adaptation with the whole dataset as the target domain.
Secondly, these datasets mainly focus on geographic diversity and do not objectively assess the diversity in scene appearances.
In addition, the analysis of existing domain adaptation methods is not the main focus of these studies, and the characteristics of different methods of domain adaptation have not been examined.
Thus, the underlying potential and technical challenges of domain adaptation have not been fully investigated with existing datasets.

This work presents \datasetname~(\datasetnamelong), a novel dataset for benchmarking domain adaptation methods on multiple diverse domains.
Our approach is to regard {\em individual road-scene videos as different target domains}.
Our dataset consists of a diverse set of individual videos (Fig.~\ref{fig:DRIV-100thumb}) and allows extensive evaluation of video-level unsupervised domain adaptation where each video individually serves as a target domain.
This evaluation setting simulates one of the most important practical application scenarios of domain adaptation, i.e., adapting segmentation models to individual cars or dashboard cameras.
In this way, this work aims at systematically analyzing the real-world potentials and challenges of domain adaptation.

To achieve this goal, our dataset consists of segmentation annotations for 100 in-the-wild YouTube videos.
To guarantee the diversity, we selected these 100 videos based on two different criteria; half of them are selected according to subjective judgment, and the other half are selected using an anomaly score estimated based on an outlier detector trained on the Cityscapes dataset~\cite{cityscapes}.
As a result, \datasetname~contains diverse road-scene videos (domains) ranging from suburban/rural areas to nature-surrounded roads.
Using \datasetname, we present a detailed performance analysis of recent representative methods of domain adaptation.
Throughout the analyses, we clarify the performance gap from an existing adaptation scenario and discuss how different scenes influence the adaptation performances.

Our major contributions are twofold. 
First, we present a novel multi-domain dataset for evaluating unsupervised domain adaptation methods of semantic segmentation in a more challenging setting with diverse target videos/domains.
We publicly release the annotation data together with target video IDs.
Second, through benchmarking analysis, we clarify the difficulties of domain adaptation methods.
Our dataset and analysis illustrate different performance characteristics from commonly-used adaptation settings and pose novel challenges for adaptation algorithms.

\section{Related work}

\subsection{Semantic Segmentation Methods and Datasets}

Semantic segmentation is the task of performing pixel-level classification on input images.
Recent advances in semantic segmentation largely depend on deep convolutional neural networks (CNNs)~\cite{FCN,segnet,PSP,U-Net,chen2017deeplab,pohlen2017full,chen2018encoder,huang2019ccnet,liu2019auto}.
Fully convolutional network (FCN)~\cite{FCN} is one of the most influential examples, and another representative approach is to use an encoder-decoder structure with a dedicated decoder to perform the segmentation task~\cite{segnet,U-Net}.
Chen et al. proposed an approach to use atrous spatial pyramid pooling to incorporate multi-scale image features and fully-connected CRFs to capture fine image details~\cite{chen2017deeplab,chen2018encoder}.
Recent research introduces the attention module to model long-range dependencies between pixels~\cite{huang2019ccnet}, and enables neural architecture search for semantic segmentation~\cite{liu2019auto}.
However, like other computer vision tasks, performances of these methods heavily depend on the choice of the training dataset.

The introduction of large-scale road-scene datasets~\cite{cityscapes,camvid,BDD100K} was one of the key driving factors in developing deep learning-based segmentation models.
For example, Cityscapes~\cite{cityscapes} is a road-scene image dataset collected in 50 cities in and around Germany, which includes 5,000 fully annotated images.
To address the difficulty of ground-truth annotation, there have been some efforts to create synthetic datasets for training semantic segmentation models~\cite{Virtual-KITTI,SYNTHIA,GTA5}.
The GTA5 dataset~\cite{GTA5} uses a game engine to capture photo-realistic road-scene images together with their segmentation labels.
However, these datasets are limited in terms of both geographical and photographic diversity of environments.
Chen et al. discussed the domain bias issue caused by existing datasets using Google Street View images, while their data is also limited to road-scene images in city areas~\cite{CrossCity}.
Mapillary Vistas is another road-scene image dataset collected from all around the world, which contains 25,000 fully annotated images~\cite{MVD2017}.
However, these datasets consisted of one-shot frames without any clear definition of domains.
As the most related to ours, Zendel et al. presented a road-scene dataset using Internet videos for assessing scene understanding methods, including semantic segmentation~\cite{zendel2018wilddash}.
However, the videos were manually selected based on a subjective judgment about hazardous factors, and most importantly, none of these works focus on a systematic analysis of domain adaptation methods.
In contrast, this work proposes an anomaly detection-based scheme for target data selection and presents detailed analyses of domain adaptation by treating individual videos as the target domain.

\subsection{Domain Adaptation for Semantic Segmentation}

Domain adaptation is a task to effectively apply a model trained on a {\em source} domain to a different {\em target} domain and has been extensively studied as both basic and applied machine learning researches.
Since it is almost impossible to obtain ground-truth labels for each target domain in practice, the standard approach for semantic segmentation is unsupervised domain adaptation.

One of the most typical approaches for domain adaptation is to make use of Generative Adversarial Networks (GANs) in addition to the base segmentation network to align the feature representations~\cite{sankaranarayanan2018learning,hong2018conditional} or input images themselves~\cite{hoffman2018cycada} between the source and target domains. 
Sankaranarayanan et al. proposed a method to combine a generator branch added to the encoder part of the base segmentation network to generate images that a discriminator can not classify as being from the source or target domain~\cite{sankaranarayanan2018learning}.

Another related approach is to apply the adversarial discriminator component directly to the final segmentation output~\cite{GAN_DA,hong2018conditional,CrossCity,luo2019taking}. 
Tsai et al. pointed out that this strategy works more effectively on semantic segmentation~\cite{GAN_DA}, and Luo et al. proposed a method to use category-level adversarial loss on the classifier outputs together with a co-training loss on two different classifiers~\cite{luo2019taking}.

Among other methods without adversarial components~\cite{zhang2017curriculum,zou2018unsupervised,zou2019confidence,saleh2018effective}, one representative approach is to generate pseudo-labels for the target domain. 
Zou et al. proposed a method to generate pseudo-labels using a model trained in the source domain while making use of only the most confident sections and iteratively increasing the proportion of the predictions used as training progresses~\cite{zou2018unsupervised,zou2019confidence}.

Most of these methods use the commonly-used synthetic-to-real domain adaptation settings (e.g., GTA5~\cite{GTA5} to Cityscapes~\cite{cityscapes}) for performance evaluation.
To provide a detailed analysis of different domain adaptation approaches on the \datasetname~dataset, we evaluate three recent methods with publicly available implementations from each of the above categories, namely LSD-Seg (Sankaranarayanan et al.~\cite{sankaranarayanan2018learning}), CLAN (Luo et al.~\cite{luo2019taking}) and CRST (Zou et al.~\cite{zou2019confidence}).

\section{\datasetname~dataset}

As briefly described earlier, our \datasetname~dataset consists of 100 YouTube videos selected based on two different criteria to guarantee the scene diversity.
For each video, we provide pixel-level segmentation annotations on four frames.
This provides the evaluation data for unsupervised domain adaptation with a comparative scale to validation/test subsets of existing datasets.
In this section, we first describe the details of the video selection and evaluation frame annotation.
We afterward present some analyses on data statistics to clarify the characteristics of \datasetname~dataset.

\subsection{Data Collection}

To establish the dataset, we first select road-scene videos from YouTube which are different from typical images in existing segmentation datasets.
The goal is to provide a set of diverse and unique road-scene videos which are expected to be challenging as domain adaptation targets.
However, it is not obvious how to judge each video's uniqueness, especially because the required property as the target domain can potentially depend on the adaptation method.

To cover contrasting sides of data selection approaches, we opted to select each half of the dataset based on different criteria.
The first half of the dataset is selected according to subjective criteria based on human judgment, and the second half is selected according to objective criteria based on anomaly scores.
This approach introduces higher diversity to the selected videos and enables an analysis of how the anomaly score is correlated with the difficulty of domain adaptation.

\subsubsection{Selection with Human Judgment}

One of our interests is picking road-scene videos captured in environments/areas that are not yet covered in the existing datasets.
We chose YouTube as the data source because it naturally covers diverse road-scene videos from all over the world.
To ensure diversity, we manually selected the first 50 videos of the \datasetname~dataset from YouTube with the following conditions.

We started by searching for videos on YouTube using query words indicating road-scene videos (such as ``dashcam'' and ``driving''), together with some additional keywords representing diverse environments (such as ``rural'', ``natural'', ``asia'', ``south-america'') in multiple languages. 
From the search results, we picked and verified videos showing diverse adaptation targets in terms of either driving condition (gravel road, natural road, highway road) or cultural/geographical background (cars driving on the left side, different appearances of buildings and plants).
All of the selected 50 videos have a resolution higher than HD ($1280 \times 720$).
We double-checked that all of these videos are captured by dashboard cameras or alike in a front-facing direction from driving cars. 
We also rejected heavily cut and edited videos to avoid any artifacts as a domain adaptation target.
If the video contains additional opening or ending shots, we trimmed the video during experiments and provide the manual annotation as part of the dataset.

\subsubsection{Selection with Anomaly Score}

While the previous criteria can guarantee that videos have some meta-properties that are different from existing road-scene datasets, it is not clear that human judgment is sufficient to create a challenging evaluation set for domain adaptation.
Therefore, we introduced another objective and systematic criteria based on an anomaly detection algorithm to select the other 50 videos.
Specifically, we employed the isolation forest algorithm~\cite{isolation} to rank candidate videos according to anomaly scores concerning the Cityscapes dataset~\cite{cityscapes}.

We first selected and downloaded 200 road-scene videos with the same restriction on video contents and resolution as above, but without any subjective judgment on recorded environments.
To judge the anomality of these videos in the context of semantic segmentation, we extracted $4096 \times H \times W$ features from the 15th convolutional layer of the FCN-8s segmentation network~\cite{FCN} trained on the Cityscapes dataset.
$W$ and $H$ can vary depending on the input image resolution, but we computed 4096-dimensional spatial mean feature vectors.
An isolation forest model was trained on feature vectors extracted from the train set of the Cityscapes dataset.
From each of the 200 YouTube videos, we extracted the same 4096-dimensional features from 2,000 frames with equal intervals.
By applying the isolation forest model to these features, we obtain anomaly scores of these frames.
Each video was represented with the median anomaly score of these 2,000 frames, and we selected the 50 videos with the lowest anomaly scores.
The implementation of the isolation forest was based on the scikit-learn library~\cite{scikit-learn} with default hyperparameters, and the lower anomaly score indicates the input is likely to be an outlier.

\subsection{Annotation Process}

\begin{figure}[t]
  \centering
  \includegraphics[width=\linewidth]{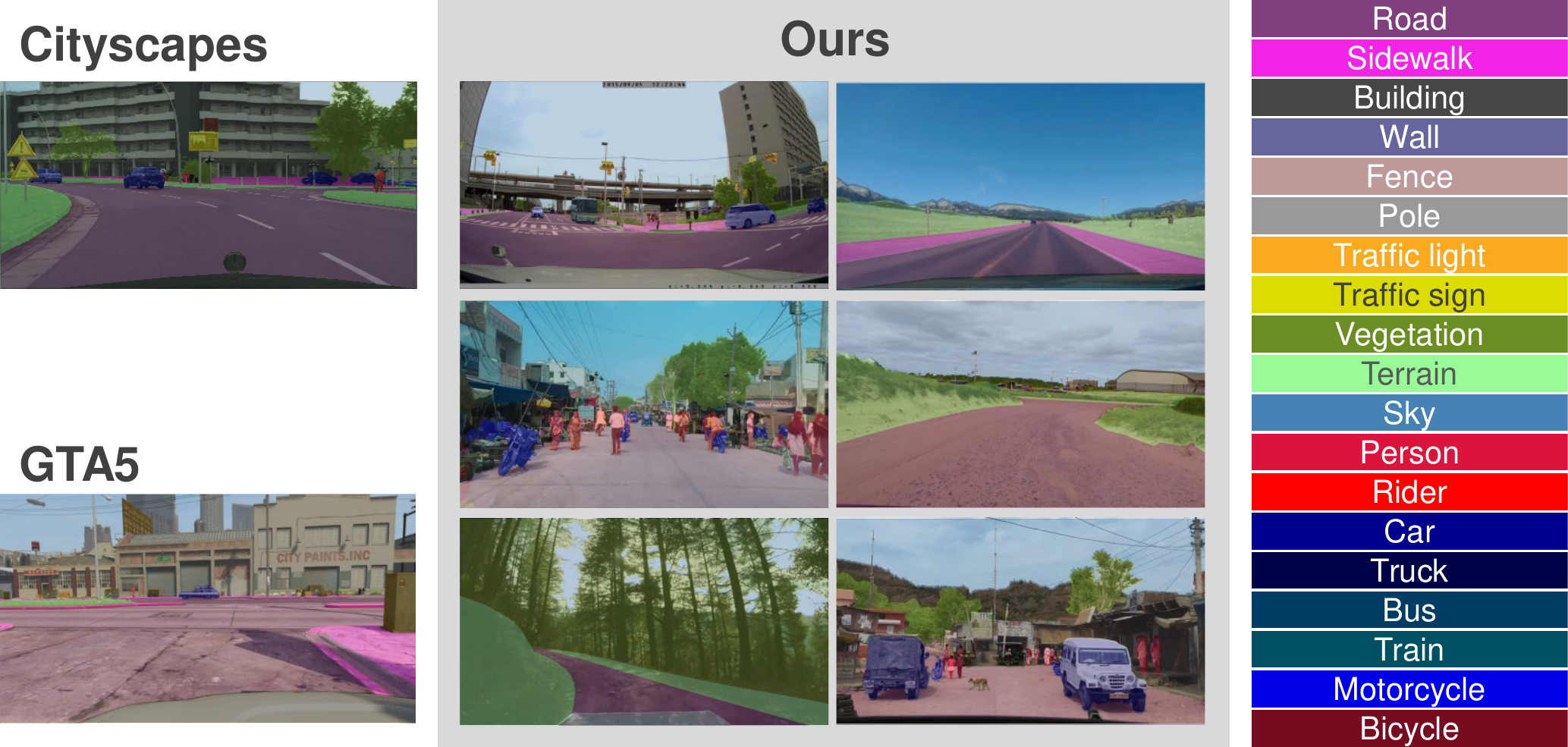} 
  \caption{Annotation examples of our dataset, Cityscapes~\cite{cityscapes}, and GTA5~\cite{GTA5}. Annotated class labels are visualized using the corresponding color shown on the right side, and overlaid on top of the original image.}
  \label{fig:DRIV-100anno}
\end{figure}

For these selected videos, we provide Cityscapes-compatible ground-truth segmentation labels on four frames per video for evaluation.
We split each video evenly into four segments and randomly select annotation-target frames from each of the four segments.
If the selected frame contains any factors making annotation difficult, such as strong motion blur and backlight, we repeated the random selection until it finds an appropriate frame.

We randomly and equally distributed these $4 \times 100$ images to 15 annotators (therefore $26.6$ images per annotator on average), and instruct them to annotate pixel-level semantic labels using an open-source tool\footnote{\url{https://github.com/Hitachi-Automotive-And-Industry-Lab/semantic-segmentation-editor}}.
All annotators were asked to familiarize themselves with the annotation tool by annotating sample images beforehand.
To maintain annotation quality and consensus among annotators, we also provided a document describing the 19 label definitions used in the Cityscapes dataset and some sample annotation results.
Annotation results were also double-checked by the authors, and re-annotated if we found any issues, mistakes, or inconsistencies.

Fig~\ref{fig:DRIV-100anno} shows some examples of annotated frames from our dataset, together with examples from the Cityscapes and GTA5 datasets.
As can be seen, our dataset provides annotations with equivalent quality and consistent class categories with existing datasets~\cite{cityscapes,GTA5}.
During annotation, we aligned resolutions of annotation target images to the minimum size and thus provide annotation results in the HD ($1280 \times 720$) format.

\subsection{Dataset Statistics}

\begin{figure}[t]
  \centering
  \includegraphics[width=\linewidth]{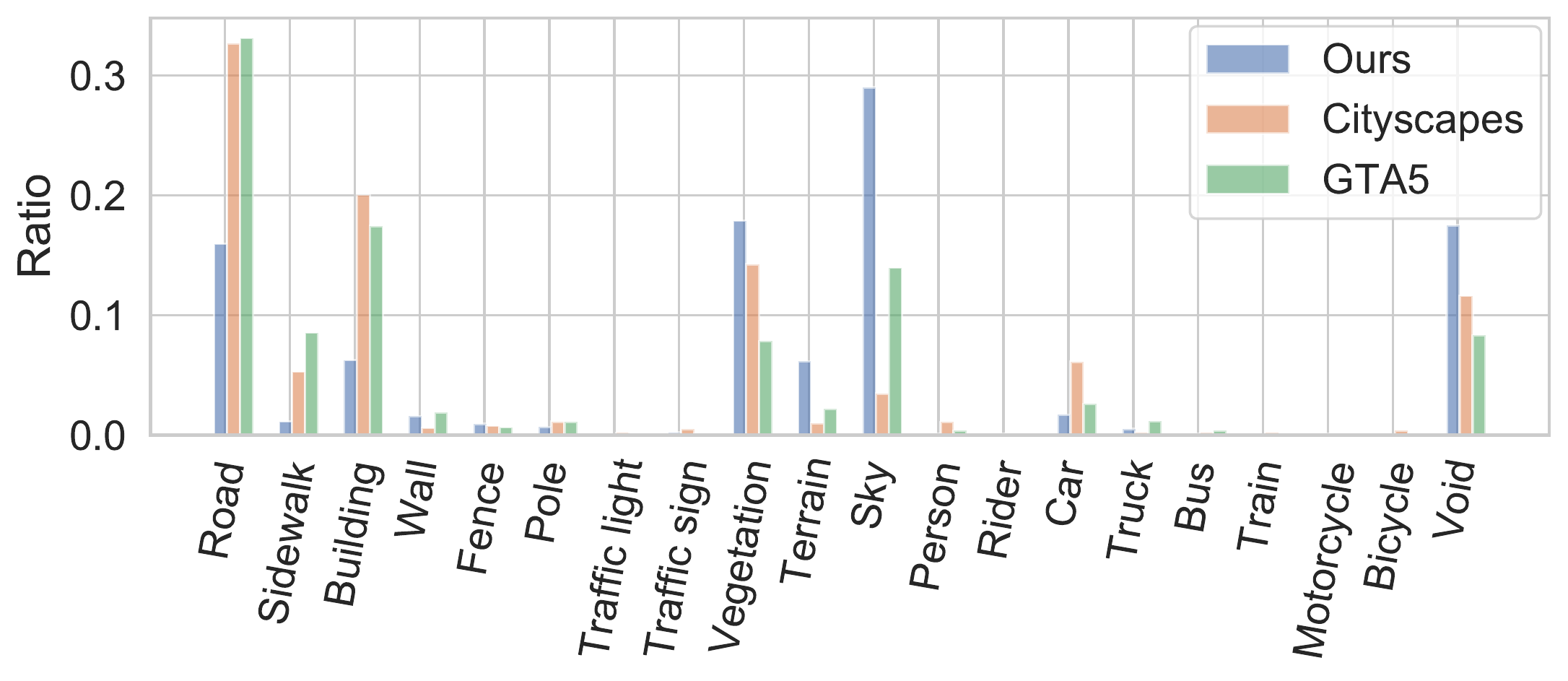} 
  \caption{Distributions of annotated pixels in our dataset, Cityscapes~\cite{cityscapes}, and GTA5~\cite{GTA5}. The horizontal axis corresponds to semantic classes, and the vertical axis shows the ratio of pixels corresponding to each class.}
  \label{fig:label_dist}
\end{figure}

Fig~\ref{fig:label_dist} shows label distributions of \datasetname, Cityscapes~\cite{cityscapes}, and GTA5~\cite{GTA5}.
The horizontal axis corresponds to the classes used in these datasets, and the vertical axis shows the ratio of pixels corresponding to each class.
Both the Cityscapes dataset and the GTA5 dataset mostly consist of city/urban scenes, and it can be seen that classes like {\em building} and {\em road} are dominant over the distributions.
In contrast, \datasetname~contains scenes with unpaved roads, and even wholly natural scenes without any human-made buildings.
As a result, \datasetname~has more pixels labelled with {\em veggetation} and {\em terrain} classes.
It can also be seen that the {\em sky} class is dominant in \datasetname, which is also because \datasetname~contains open areas without any tall buildings.

\section{Experiments}

In this section, we provide analyses of several previously proposed domain adaptation methods using the \datasetname~dataset. 
Specifically, we compare the GTA5-to-\datasetname~adaptation (GTA5~\cite{GTA5} as source domain and \datasetname~as target domain) performance with the community-standard GTA5-to-Cityscapes adaptation (GTA5 as the source domain and Cityscapes~\cite{cityscapes} as target domain) performance to clarify the uniqueness of our dataset as the adaptation target.

\subsection{Evaluation Protocol}

\subsubsection{Evaluation Metrics}

\begin{table*}[t]
\caption{Summary of domain adaptation performances. Each column shows mIoU scores of each combination of base network and adaptation method with the GTA5 dataset as the source domain. The best performances are highlighted in bold.}
\label{tab:adaptation_to_cityscapes_results}
\centering
\resizebox{0.7\linewidth}{!}{
{\def\arraystretch{1.1}\tabcolsep=7pt

\begin{tabular}{l|c|cccc|ccc}
\hline

Target & Metric & \multicolumn{4}{c|}{FCN} & \multicolumn{3}{c}{DeepLab} \\ \hline
 & & Unadapted & LSD-Seg & CLAN & CRST & Unadapted & CLAN & CRST \\
\datasetname & Class-wise Avg. mIoU & 15.2 & 16.7 & \textbf{20.9} & 15.3 & 19.2 & \textbf{25.1} & 22.0 \\
\datasetname & Video-wise Avg. mIoU & 16.8 & 18.0 & \textbf{26.6} & 16.8 & 21.5 & \textbf{29.3} & 24.5 \\ \hline

Cityscapes & mIoU & 25.3 & \textbf{34.7} & 28.7 & 31.8 & 34.1 & 38.7 & \textbf{43.5} \\ \hline

\end{tabular}
}}
\end{table*}

Following many prior works~\cite{FCN,segnet,PSP,CrossCity,pascol_voc,PASCAL_CONTEXT}, we adapt mean Intersection over Union (mIoU) metric~\cite{pascol_voc} as a performance indicator, which is defined as
\begin{equation}
    \textrm{mIoU} = \frac{1}{K}\sum_{k}^{K}\frac{A_k \cap B_k}{A_k \cup B_k},
\end{equation}
where $A_k$ is the set of pixels with the ground-truth labels of $k$-th class, and $B_k$ is the one in the segmentation result.
$K$ is the total number of classes, and mIoU is defined as the mean of IoU scores of $K$ classes.

In \datasetname, each video represents a unique domain.
The label distribution significantly differs across videos, and all of the $K=19$ classes do not always appear in each video.
Therefore, when calculating mIoU, we ignore any classes that do not appear in either ground-truth or predicted labels.
In this way, perfect predictions result in $100\%$ mIoU even if not all of the classes are present, while the score still penalizes false predictions of non-existent classes.

To properly reflect this dataset characteristic containing minor classes, we report both class-wise average mIoU and video-wise average mIoU in the experiments.
The video-wise average mIoU is defined as the average of mIoU scores of each video (with different $K$s).
In contrast, the class-wise average mIoU is defined as the average over all $K=19$ classes using the mean of the IoU scores of each individual class in all $100$ videos.
These two metrics are not exactly the same, and in \datasetname~the video-wise average mIoU tends to perform better due to the presence of rare classes like {\em train}.

\subsubsection{Datasets}

As discussed above, \datasetname~dataset consists of videos whose minimum resolution is $1280 \times 720$ pixels. 
The total lengths of these videos range from $2,329$ to $153,799$ frames.
To align the number of unique frames used during the adaptation for each video, $2,329$ frames were randomly selected from each video as the unsupervised domain adaptation target.
The Cityscapes dataset is composed of $5,000$ images at $2048 \times 1024$ resolution, and we used the $2,975$ training subset for training and $500$ validation subset for evaluation.
The GTA5 dataset consists of $24,966$ synthetic images mostly at $1914 \times 1052$ resolution.
Since IoU-based evaluation metrics can also depend on the target image resolution, we aligned the image resolution to $1280 \times 720$ throughout the adaptation experiments. 
For images larger than $1280 \times 720$ pixels, we first resized the image so that the height becomes 720 pixels, and then center-cropped to match the $1280 \times 720$ resolution.

\subsection{Domain Adaptation Methods}

We employ two base segmentation networks in evaluation: FCN (FCN-8s architecture~\cite{FCN} with the VGG-16~\cite{VGG} backbone) and DeepLab (DeepLab-v2 architecture~\cite{chen2017deeplab} with the Resnet-101~\cite{resnet} backbone).
As domain adaptation methods, we employ three state-of-the-art methods: LSD-Seg~\cite{sankaranarayanan2018learning}, CLAN~\cite{luo2019taking}, and CRST~\cite{zou2019confidence}.
FCN is evaluated with all three adaptation methods, while DeepLab is evaluated with CLAN and CRST.

\subsubsection{Implementation Details}
For all of the three methods, we used the official code published by the authors.
During the adaptation, we use the resized GTA5 dataset as the source domain and the frames of the respective video from \datasetname~as the target domain.
Following the description in their original papers, we use the patch size of $1024 \times 512$ for all three methods.

Since there is no validation data available in \datasetname~(as most of the unsupervised domain adaptation settings in practice), testing is done using the models saved after the full training process defined in the author implementations.
Although this does not guarantee the best adaptation performance in all of the 100 videos, the average mIoU scores still effectively represent the model performances in practical use cases.

\paragraph{LSD-Seg.}
The training procedure described for LSD-Seg consists of $100,000$ training iterations, with the backbone CNNs pre-trained on ImageNet~\cite{imagenet} and the rest of the weights trained from scratch.
Training is done using $1024 \times 512$ random crops, and the final evaluation is done directly on the $1280 \times 720$ \datasetname~images.
The authors only reported on and provided implementation for FCN.

\paragraph{CLAN.}
The training procedure for CLAN also consists of $100,000$ training iterations, using backbone CNNs pre-trained on ImageNet with the rest of the weights trained from scratch.
The authors only provided an implementation for DeepLab while also reporting results with FCN in the paper. 
For better comparison, therefore, we also modified the code to use FCN.
The patch size is still $1024 \times 512$ for training with FCN, but the input images are resized instead of cropped.
Therefore, during the evaluation, we also resize the input images to $1024 \times 512$ and then upscale the output to the full $1280 \times 720$.

\paragraph{CRST.}
Unlike the previous two methods, CRST adapts a model that was already trained on the source domain.
The authors provided an implementation for DeepLab together with pre-trained weights on the GTA5 dataset.
For FCN, we pre-trained the network on GTA5 by following the details described in the paper.
The self-training process described by the authors consists of $3$ rounds with $2$ epochs each.
Following the original implementation using a subset of $505$ images from the target domain (assuming Cityscapes), we also used a random subset of $505$ frames from the initial selection of $2,329$ frames in \datasetname. 
While various regularization methods were introduced in \cite{zou2019confidence}, we use MRKLD, which performed the best in their evaluation. 

\subsection{Evaluation of Domain Adaptation}

In Table~\ref{tab:adaptation_to_cityscapes_results}, we first summarize the domain adaptation results of all methods.
Each column shows mIoUs of each combination of base network and adaptation method with the GTA5 dataset as the source domain.
The first two rows show adaptation results on our \datasetname~dataset (both class-wise and video-wise averages), while the last row shows adaptation results on the Cityscapes dataset.
In the commonly-used GTA5-to-Cityscapes setting, LSD-Seg and CRST performed the best with FCN and DeepLab, respectively.
However, LSD-Seg showed only minor improvements with FCN in the GTA5-to-\datasetname~setting.
CRST showed some improvement with DeepLab, but performs poorly with FCN.
In contrast, CLAN achieved the best performance with both of the segmentation networks.

\begin{table*}[t]
\caption{Details of adaptation results to \datasetname, with individual IoU scores of each class. The second column indicates the base segmentation network architecture (F: FCN, D: DeepLab). The performances of an unadapted model trained on Cityscapes are also included for reference. The best performances are highlighted in bold.}
\label{tab:adaptation_to_driv_results}
\resizebox{\textwidth}{!}{
{\def\arraystretch{1.1}\tabcolsep=7pt

\begin{tabular}{l|c|ccccccccccccccccccc|cc}
\hline

 & \rotatebox{90}{Arch.} & \rotatebox{90}{road} & \rotatebox{90}{sidewalk} & \rotatebox{90}{building} & \rotatebox{90}{wall} & \rotatebox{90}{fence} & \rotatebox{90}{pole} & \rotatebox{90}{light} & \rotatebox{90}{sign} & \rotatebox{90}{vegetation} & \rotatebox{90}{terrain} & \rotatebox{90}{sky} & \rotatebox{90}{person} & \rotatebox{90}{rider} & \rotatebox{90}{car} & \rotatebox{90}{truck} & \rotatebox{90}{bus} & \rotatebox{90}{train} & \rotatebox{90}{motorcycle} & \rotatebox{90}{bicycle} & \rotatebox{90}{\parbox{2cm}{class-wise \\ average mIoU}} & \rotatebox{90}{\parbox{2cm}{video-wise \\ average mIoU}} \\ \hline
\multicolumn{23}{c}{Source: GTA5, Target: \datasetname} \\ \hline
Unadapted & F & 17.7 & 5.1 & 16.8 & 6.2 & 8.8 & 19.3 & 6.3 & 16.7 & 49.8 & 19.8 & 58.8 & \textbf{12.9} & 9.6 & 20.7 & 5.0 & 0.9 & 0.0 & \textbf{10.0} & 3.8 & 15.2 & 16.8 \\
LSD Seg & F & 41.3 & 5.5 & 22.4 & 8.3 & 10.2 & 21.2 & 4.3 & 14.0 & 53.6 & 16.6 & 61.9 & 6.8 & 5.6 & 29.1 & 6.8 & 1.6 & 0.0 & 4.2 & 3.3 & 16.7 & 18.0 \\
CLAN & F & \textbf{53.0} & 5.8 & \textbf{29.2} & \textbf{8.8} & 10.2 & 18.4 & 6.7 & 15.4 & \textbf{65.1} & \textbf{27.5} & \textbf{90.0} & 8.7 & 4.7 & \textbf{35.9} & \textbf{10.8} & \textbf{3.9} & 0.0 & 1.2 & 0.9 & \textbf{20.9} & \textbf{26.6} \\
CRST & F & 21.1 & \textbf{6.2} & 18.5 & 7.2 & \textbf{10.7} & \textbf{22.2} & \textbf{7.3} & \textbf{18.5} & 42.1 & 19.7 & 50.9 & 9.6 & \textbf{9.9} & 27.6 & 4.7 & 0.0 & \textbf{0.1} & 9.0 & \textbf{5.0} & 15.3 & 16.8 \\ \hline
Unadapted & D & 38.3 & 5.9 & 27.1 & 9.2 & 10.3 & 20.3 & \textbf{7.3} & 19.0 & 55.6 & 23.4 & 80.2 & 9.7 & \textbf{12.2} & 19.2 & 7.3 & 2.7 & 0.0 & \textbf{11.3} & 6.2 & 19.2 & 21.5 \\
CLAN & D & \textbf{68.2} & 9.7 & \textbf{33.7} & 13.7 & 9.6 & 18.8 & 6.8 & 21.4 & \textbf{65.8} & \textbf{31.5} & \textbf{89.3} & 10.2 & 10.7 & \textbf{47.1} & \textbf{16.9} & \textbf{7.3} & 0.0 & 10.4 & 6.0 & \textbf{25.1} & \textbf{29.3} \\
CRST & D & 45.6 & \textbf{10.8} & 31.5 & \textbf{13.9} & \textbf{14.5} & \textbf{26.4} & 7.2 & \textbf{24.0} & 49.5 & 27.5 & 79.2 & \textbf{10.4} & 10.2 & 30.2 & 11.4 & 4.7 & 0.0 & 10.9 & \textbf{9.4} & 22.0 & 24.5 \\ \hline
\multicolumn{23}{c}{Source: Cityscapes} \\ \hline
Unadapted & F & 47.9 & 10.9 & 25.1 & 7.2 & 14.8 & 21.2 & 4.2 & 24.7 & 56.3 & 25.2 & 91.0 & 5.9 & 8.9 & 22.5 & 8.5 & 1.8 & 0.0 & 4.9 & 5.1 & 20.3 & 21.8 \\
Unadapted & D & 73.9 & 18.0 & 30.9 & 12.4 & 16.4 & 25.1 & 8.1 & 27.4 & 60.8 & 30.8 & 86.8 & 6.6 & 11.7 & 41.4 & 14.4 & 5.0 & 0.2 & 4.1 & 6.6 & 25.3 & 27.7 \\ \hline

\end{tabular}
}}

\end{table*}

In Table~\ref{tab:adaptation_to_driv_results} we further show the details of adaptation results to \datasetname, with individual IoU scores of each class.
We also show the performances of unadapted segmentation networks trained on Cityscapes as a reference.
With both of the segmentation networks, we see a strong tendency for CLAN to perform best on large classes like {\em road} and {\em vegetation}, whereas CRST achieves slightly better scores on small classes such as {\em pole} or {\em sign}. 
This indicates that CRST's class-balanced training strategy encouraged the inclusion of less confident classes, and some miss-attribution of large classes led to the lower scores on average. 
On the other hand, LSD-Seg does not achieve the best performance for any classes and achieve mediocre performance between the other two methods. 

\begin{figure}[t]
  \centering
  \includegraphics[width=\linewidth]{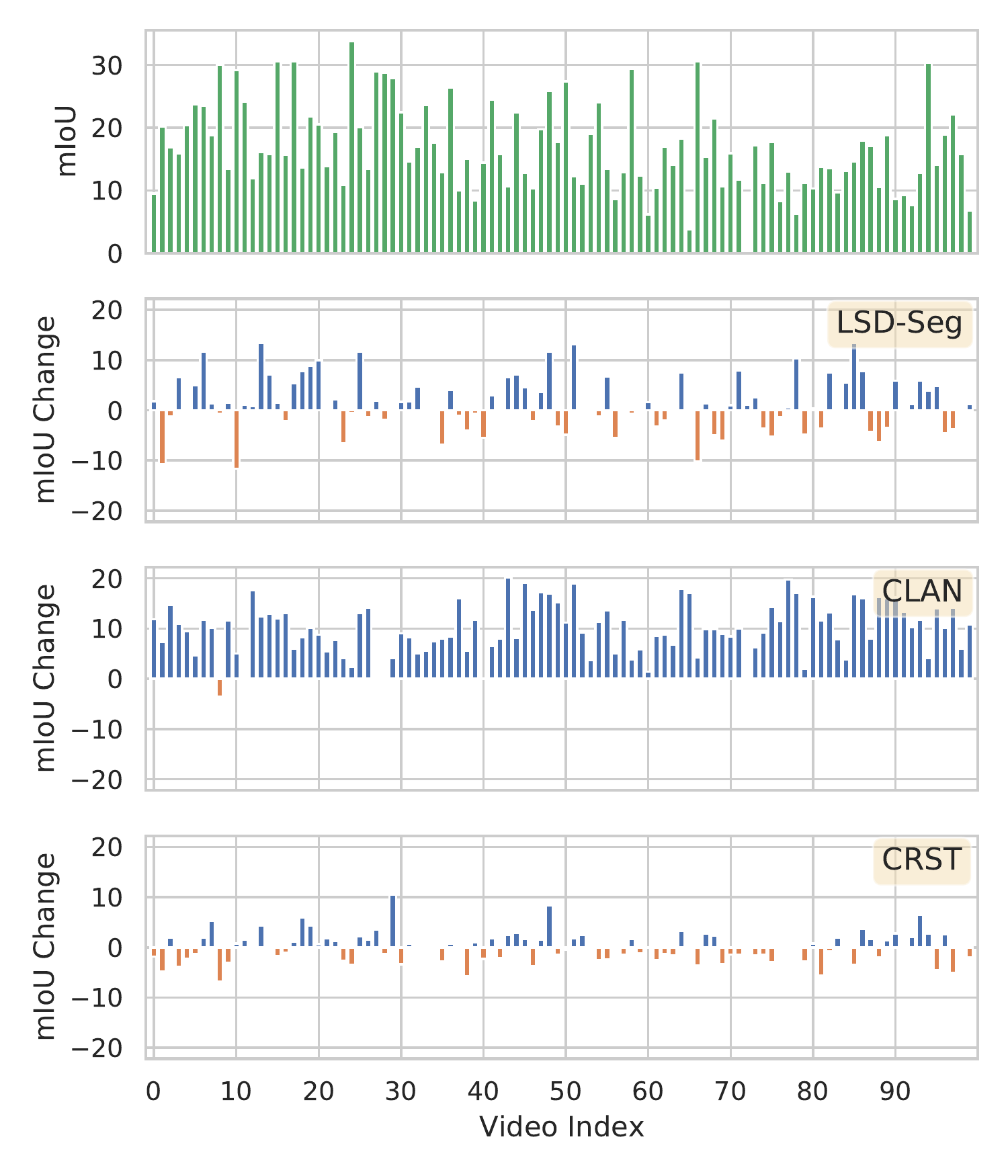} 
  \caption{Details of per-video adaptation performance with FCN. The top graph shows the unadapted mIoUs, and the other graphs show the changes in mIoU scores using each of the adaptation methods.}
  \label{fig:adapted_relative_vgg16fcn}
\end{figure}

\begin{figure}[t]
  \centering
  \includegraphics[width=\linewidth]{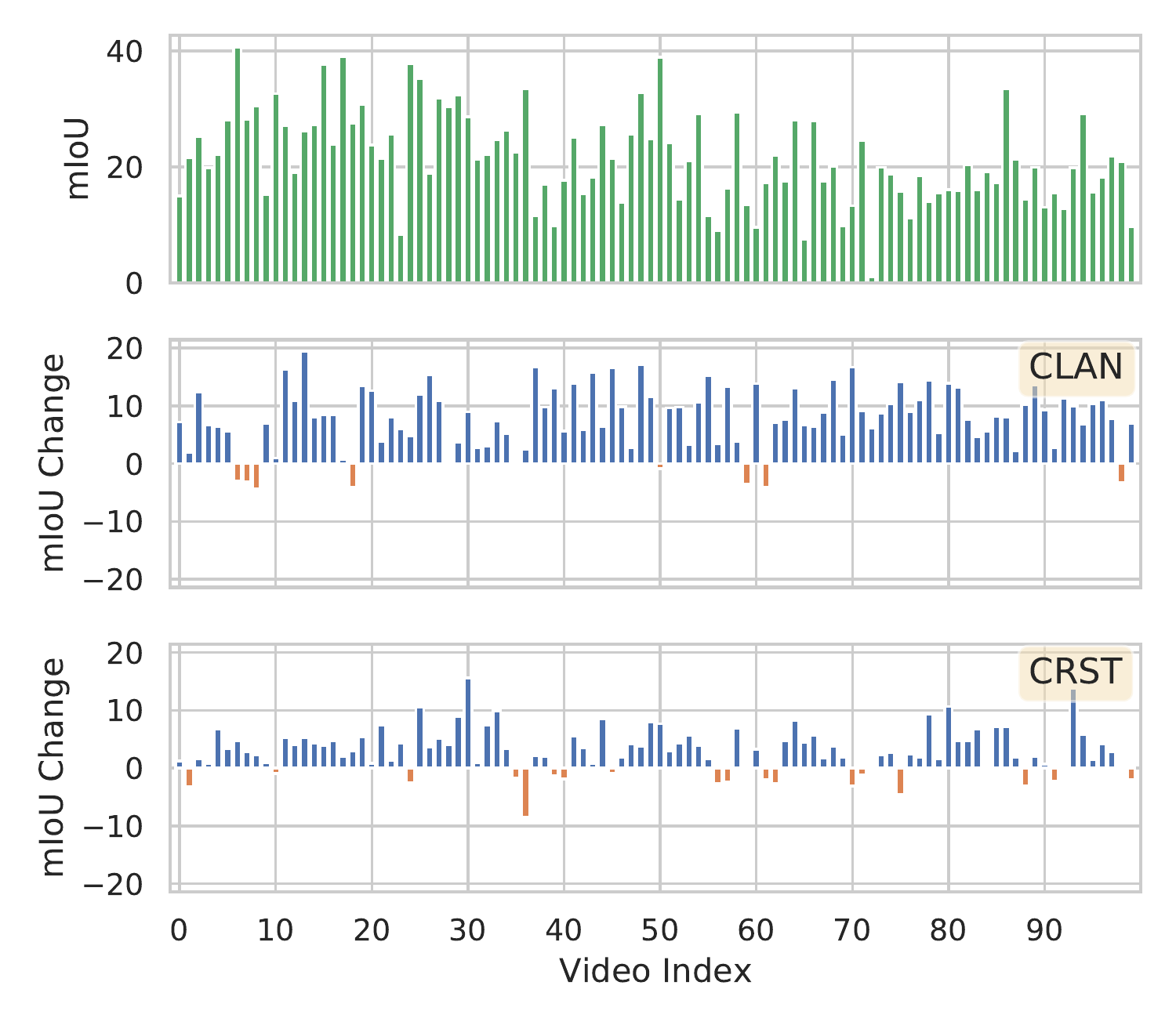} 
  \caption{Details of per-video adaptation performance with DeepLab. The top graph shows the unadapted mIoUs, and the other graphs show the changes in mIoU scores using each of the adaptation methods.}
  \label{fig:adapted_relative_deeplab}
\end{figure}

Fig~\ref{fig:adapted_relative_vgg16fcn} and \ref{fig:adapted_relative_deeplab} show visualizations of the per-video performance.
The top graph shows the unadapted mIoUs of each segmentation network (FCN in Fig~\ref{fig:adapted_relative_vgg16fcn} and DeepLab in Fig.~\ref{fig:adapted_relative_deeplab}), with the horizontal axis indicating the video indices in \datasetname.
The first 50 videos correspond to the ones selected with human judgment, and the last 50 correspond to the anomaly score-based selection.
The other graphs show the changes in mIoUs using each of the adaptation methods.

In Fig~\ref{fig:adapted_relative_vgg16fcn}, although LSD-Seg increases the performance slightly on average, the performance varies widely across videos, with some cases even performing significantly worse than the unadapted baseline, indicating a somewhat unstable adaptation process.
In contrast, CLAN shows overall positive improvements in both Fig~\ref{fig:adapted_relative_vgg16fcn} and Fig~\ref{fig:adapted_relative_deeplab}.
It shows a significant increase on average, and the decrease is only minor in the few cases where the performance drops.
From the CRST performances, we can also see that the change in performance does not directly depend on the initial mIoU score of the target video, despite the accuracy of the generated pseudo-labels being related to it. 
One possible reason is that the higher the initial mIoU is, the more difficult it is to increase the score, which acts as an opposing force to the more accurate pseudo labels.

\subsection{Qualitative Analysis}

\begin{figure}[t]
  \centering
  \includegraphics[width=\linewidth]{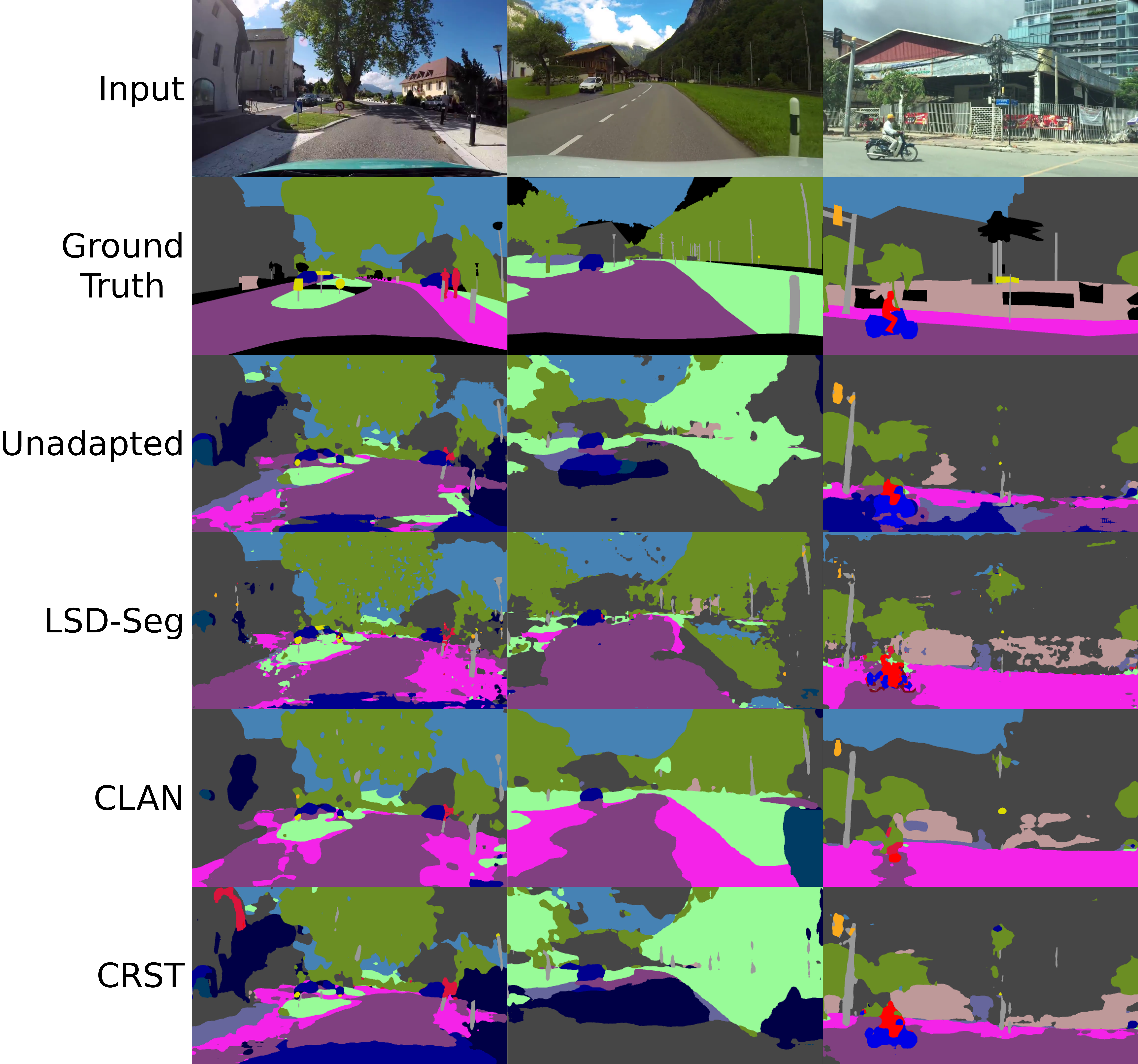} 
  \caption{Visualizations of the segmentation results with FCN. From left to right, each column shows frames from videos $13$, $43$ and $29$.}.
  \label{fig:vgg16fcn_segmentations}
\end{figure}

\begin{figure}[t]
  \centering
  \includegraphics[width=\linewidth]{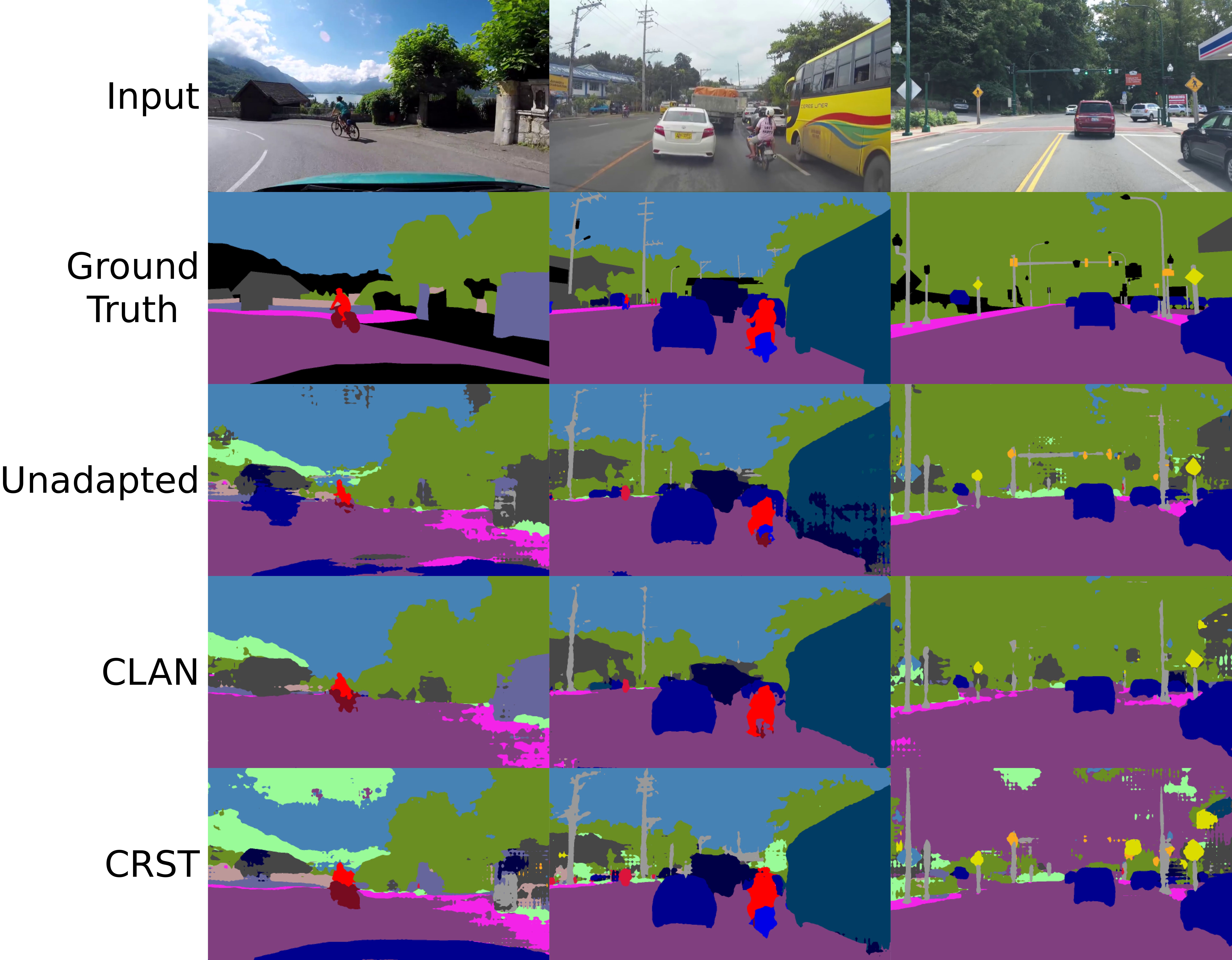} 
  \caption{Visualizations of the segmentation results with DeepLab. From left to right, each column shows frames from videos $13$, $30$ and $36$.}.
  \label{fig:deeplab_segmentations}
\end{figure}

Fig~\ref{fig:vgg16fcn_segmentations} shows some examples of segmentation results using FCN.
The columns correspond to the adaptations with the most significant improvement for LSD-Seg (video $13$), CLAN (video $43$), and CRST (video $29$), respectively.
LSD-Seg showed some improvements in {\em road} and {\em sidewalk}; however, the segmentation output still has many artifacts and some unreasonable predictions such as the {\em sky} region at the road level in the second video.
CLAN tends to produce smoother segmentation results and shows particularly good performance with the {\em sky}, {\em vegetation}, and  {\em terrain} classes.
However, unlike other methods, it completely missed the {\em motorcycle} region in the third video.
Compared to these two methods, CRST resulted in more noisy segmentation results.
It can be seen that minor artifacts in the unadapted model are amplified through adaptation, such as the {\em building} regions in the sky.
However, CRST successfully improved the segmentation in the last video and was the only method that dealt well with the {\em road} and {\em motorcycle} regions in this scenario.

Similarly, Fig~\ref{fig:deeplab_segmentations} shows some examples of the segmentation results using DeepLab. 
The first two columns correspond to the adaptation results with the most significant improvement for CLAN (video $13$) and CRST  (video $30$), respectively, and the last one shows a severe failure case of CRST (video $36$). 
Although there are still some missing details such as the {\em motorcycle} in the second video or the {\em traffic light} regions in the third video, CLAN performs reasonably in most of the cases.
From the CRST results, we can see a weakness of using pseudo-labels for adaptation.
While the initial unadapted segmentation performs relatively well on the third video, most of the top half region was miss-classified as road after adaptation.

\begin{figure}[t]
  \centering
  \includegraphics[width=\linewidth]{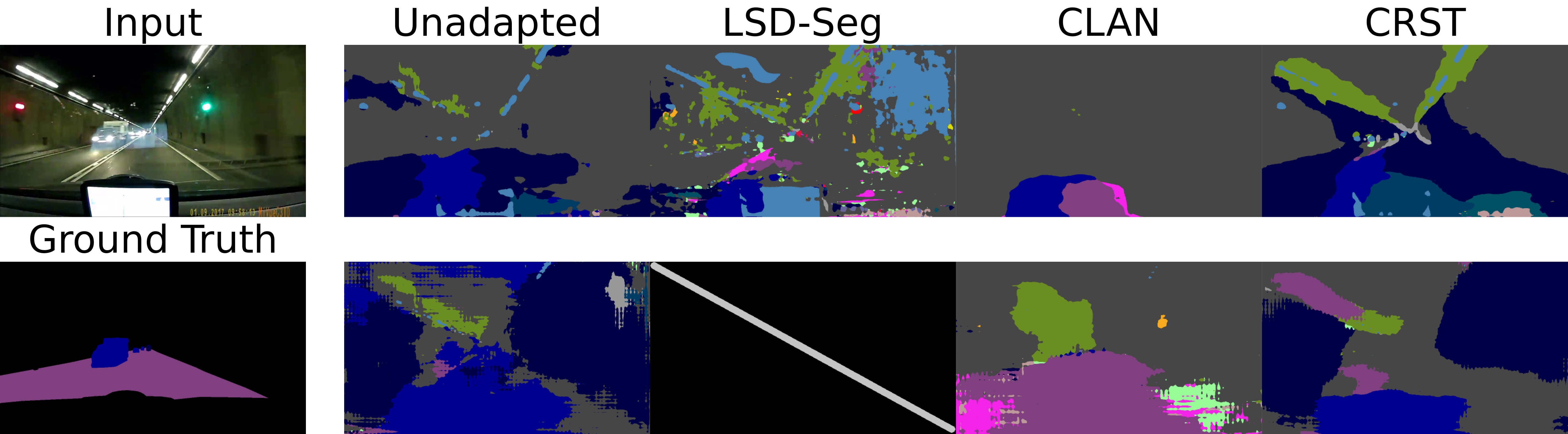} 
  \caption{Visualizations of segmentation results on one of the most challenging cases (video $72$) in our dataset. On the right side, segmentations in the top row were produced using FCN and those on the bottom using DeepLab.}
  \label{fig:vid_72}
\end{figure}

Fig.~\ref{fig:vid_72} shows one of the most challenging cases (video $72$) in our dataset, where the video is taken while driving in a tunnel with only a few classes present.
The initial mIoU is close to $0$ regardless of the segmentation network. 
Only CLAN with DeepLab achieved some minor improvement, however, qualitatively the performance is still unsatisfactory.
The oncoming vehicles are almost completely ignored in all of the cases.

\subsection{Evaluation of Data Selection Strategy}

\begin{figure}[t]
  \centering
  \includegraphics[width=\linewidth]{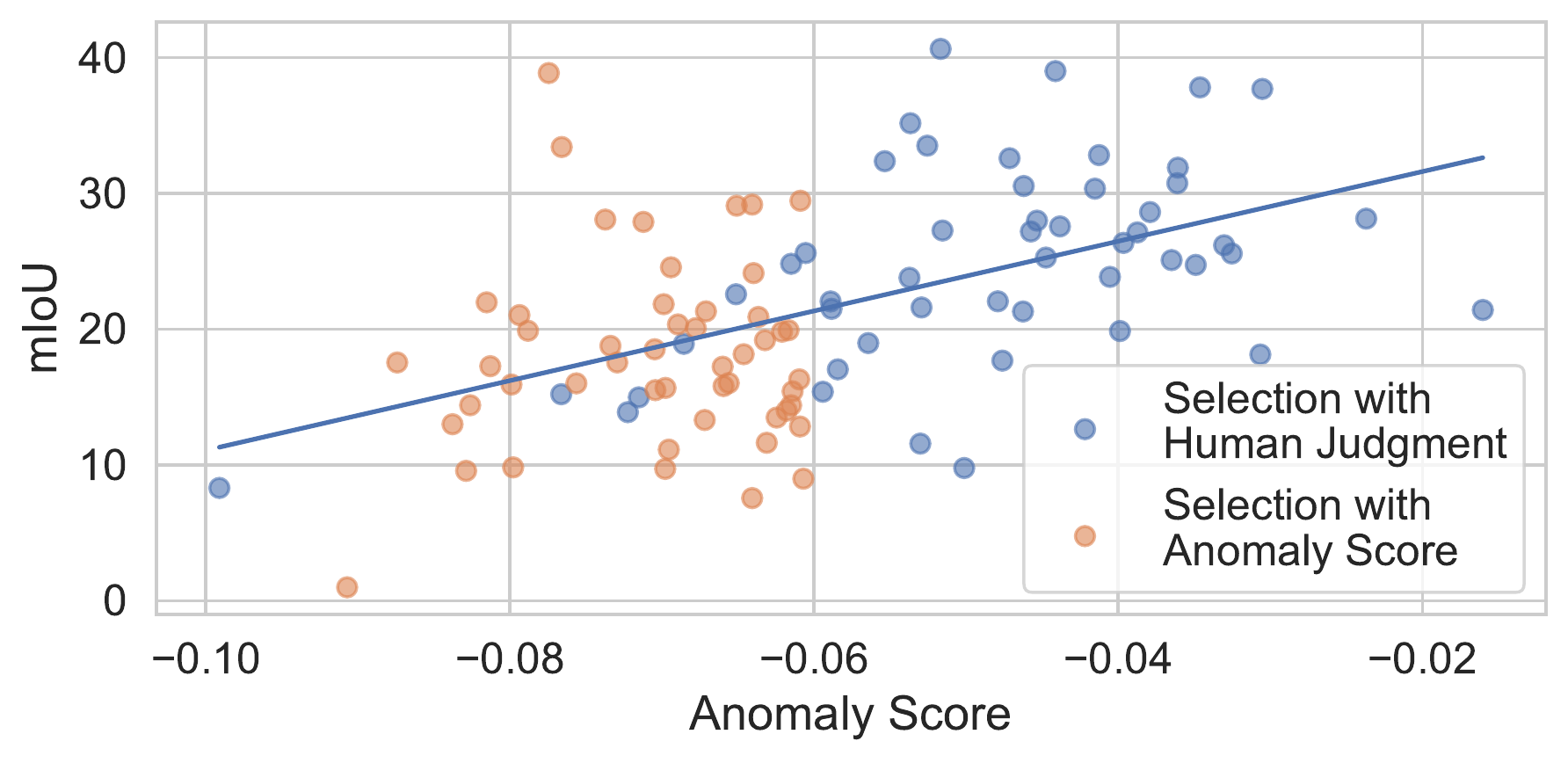} 
  \caption{The relationship between anomaly scores and unadapted mIoU scores of videos in our dataset. The horizontal axis indicates anomaly scores (lower value corresponds to higher anomality), and the vertical axis indicates mIoU scores from an unadapted DeepLab segmentation model trained on the GTA5 dataset.}
  \label{fig:selection_val_GTA}
\end{figure}

In this section, we provide an additional analysis of our data collection strategy using anomaly scores.
Fig~\ref{fig:selection_val_GTA} visualizes the relationship between anomaly scores and unadapted mIoU scores of the 100 videos in our dataset.
The horizontal axis indicates anomaly scores, and a lower value corresponds to higher anomality.
The vertical axis indicates mIoU scores of the unadapted DeepLab model trained on the GTA5 dataset.
There was a moderate correlation between anomaly scores and mIoU, with a correlation coefficient of $0.5171$.
Videos selected with anomaly detection naturally have lower anomaly scores, and thus lower mIoU scores on average.

On the first $50$ videos selected with human judgment, DeepLab trained on GTA5 reached a video-wise average mIoU of $24.9$.
In contrast, the video-wise average was only $18.2$ on the last $50$ videos selected with anomaly scores. 
However, the adaptation performance did not show such a clear cut difference between the two subsets.
CRST showed a slightly bigger improvement on the first $50$ videos, whereas CLAN had a slightly bigger improvement on the last $50$ videos.
This indicates that, while the anomaly score can be a good predictor for the difficulty of semantic segmentation, the difficulty of domain adaptation also depends on the adaptation method and other scene characteristics.
In this way, our two data selection strategies complement each other and provide more diverse test cases for domain adaptation.

\section{Conclusion}

We presented a new dataset, \datasetname, for evaluating unsupervised domain adaptation methods of road-scene semantic segmentation on multiple diverse domains.
We created our dataset by providing manual ground-truth annotations on 100 YouTube videos selected based on two different criteria, subjective human judgment and an objective anomaly score.
Using the dataset, we analyzed the performances of multiple state-of-the-art domain adaptation methods on the video-level adaptation task where each of the 100 videos serve as an individual adaptation target domain.
By comparing their performances with the commonly-used adaptation setting, our study highlighted novel challenges and difficulties of domain adaptation.

{\small
\bibliographystyle{ieee_fullname}
\bibliography{egbib}

\begin{thebibliography}{10}\itemsep=-1pt

\bibitem{segnet}
Vijay Badrinarayanan, Alex Kendall, and Roberto Cipolla.
\newblock Segnet: {A} deep convolutional encoder-decoder architecture for image
  segmentation.
\newblock In {\em IEEE Transactions on Pattern Analysis and Machine
  Intelligence}, pages 2481--2495, 2017.

\bibitem{camvid}
Gabriel~J. Brostow, Julien Fauqueur, and Roberto Cipolla.
\newblock Semantic object classes in video: A high-definition ground truth
  database.
\newblock {\em Pattern Recognition Letters}, pages 88--97, 2009.

\bibitem{chen2017deeplab}
Liang-Chieh Chen, George Papandreou, Iasonas Kokkinos, Kevin Murphy, and Alan~L
  Yuille.
\newblock Deeplab: Semantic image segmentation with deep convolutional nets,
  atrous convolution, and fully connected crfs.
\newblock {\em IEEE Transactions on pattern analysis and machine intelligence},
  40(4):834--848, 2017.

\bibitem{chen2018encoder}
Liang-Chieh Chen, Yukun Zhu, George Papandreou, Florian Schroff, and Hartwig
  Adam.
\newblock Encoder-decoder with atrous separable convolution for semantic image
  segmentation.
\newblock In {\em Proc. European Conference on Computer Vision (ECCV)}, pages
  801--818, 2018.

\bibitem{CrossCity}
Yi-Hsin Chen, Wei-Yu Chen, Yu-Ting Chen, Bo-Cheng Tsai, Yu-Chiang~Frank Wang,
  and Min Sun.
\newblock No more discrimination: Cross city adaptation of road scene
  segmenters.
\newblock In {\em Proc. International Conference on Computer Vision (ICCV)},
  pages 1992--2001, 2017.

\bibitem{cityscapes}
Marius Cordts, Mohamed Omran, Sebastian Ramos, Timo Rehfeld, Markus Enzweiler,
  Rodrigo Benenson, Uwe Franke, Stefan Roth, and Bernt Schiele.
\newblock The cityscapes dataset for semantic urban scene understanding.
\newblock In {\em Proc. Computer Vision and Pattern Recognition (CVPR)}, pages
  3213--3223, 2016.

\bibitem{imagenet}
Jia Deng, Wei Dong, Richard Socher, Li-Jia Li, Kai Li, and Li Fei-Fei.
\newblock Imagenet: A large-scale hierarchical image database.
\newblock In {\em Proc. Computer Vision and Pattern Recognition (CVPR)}, pages
  248--255, 2009.

\bibitem{pascol_voc}
Mark Everingham, Van Gool, Christopher~KI Williams, John Winn, and Andrew
  Zisserman.
\newblock The pascal visual object classes (voc) challenge.
\newblock {\em International Journal of Computer Vision}, page 303–338, 2010.

\bibitem{Virtual-KITTI}
A Gaidon, Q Wang, Y Cabon, and E Vig.
\newblock Virtual worlds as proxy for multi-object tracking analysis.
\newblock In {\em Proc. Computer Vision and Pattern Recognition (CVPR)}, 2016.

\bibitem{kitti}
Andreas Geiger, Philip Lenz, Christoph Stiller, and Raquel Urtasun.
\newblock Vision meets robotics: The kitti dataset.
\newblock {\em The International Journal of Robotics Research},
  32(11):1231--1237, 2013.

\bibitem{resnet}
Kaiming He, Xiangyu Zhang, Shaoqing Ren, and Jian Sun.
\newblock Deep residual learning for image recognition.
\newblock In {\em Proc. Computer Vision and Pattern Recognition (CVPR)}, pages
  770--778, 2016.

\bibitem{HernandezBMVC17}
Daniel Hernandez-Juarez, Lukas Schneider, Antonio Espinosa, David Vazquez,
  Antonio~M. Lopez, Uwe Franke, Marc Pollefeys, and Juan~Carlos Moure.
\newblock Slanted stixels: Representing san francisco’s steepest streets.
\newblock In {\em Proc. British Machine Vision Conference (BMVC)}, 2017.

\bibitem{hoffman2018cycada}
Judy Hoffman, Eric Tzeng, Taesung Park, Jun-Yan Zhu, Phillip Isola, Kate
  Saenko, Alexei Efros, and Trevor Darrell.
\newblock Cycada: Cycle-consistent adversarial domain adaptation.
\newblock In {\em Proc. International Conference on Machine Learning (ICML)},
  pages 1989--1998, 2018.

\bibitem{hong2018conditional}
Weixiang Hong, Zhenzhen Wang, Ming Yang, and Junsong Yuan.
\newblock Conditional generative adversarial network for structured domain
  adaptation.
\newblock In {\em Proc. Computer Vision and Pattern Recognition (CVPR)}, pages
  1335--1344, 2018.

\bibitem{huang2019ccnet}
Zilong Huang, Xinggang Wang, Lichao Huang, Chang Huang, Yunchao Wei, and Wenyu
  Liu.
\newblock Ccnet: Criss-cross attention for semantic segmentation.
\newblock In {\em Proc. International Conference on Computer Vision (ICCV)},
  pages 603--612, 2019.

\bibitem{FCN}
Long Jonathan, Shelhamer Evan, and Darrell Trevor.
\newblock Fully convolutional networks for semantic segmentation.
\newblock In {\em Proc. Computer Vision and Pattern Recognition (CVPR)}, pages
  3431--3440, 2015.

\bibitem{liu2019auto}
Chenxi Liu, Liang-Chieh Chen, Florian Schroff, Hartwig Adam, Wei Hua, Alan~L
  Yuille, and Li Fei-Fei.
\newblock Auto-deeplab: Hierarchical neural architecture search for semantic
  image segmentation.
\newblock In {\em Proc. Computer Vision and Pattern Recognition (CVPR)}, pages
  82--92, 2019.

\bibitem{isolation}
Fei~Tony Liu, Kai~Ming Ting, and Zhi-Hua Zhou.
\newblock Isolation forest.
\newblock In {\em Proc. International Conference on Data Mining (ICDM)}, pages
  413--422. IEEE, 2008.

\bibitem{luo2019taking}
Yawei Luo, Liang Zheng, Tao Guan, Junqing Yu, and Yi Yang.
\newblock Taking a closer look at domain shift: Category-level adversaries for
  semantics consistent domain adaptation.
\newblock In {\em Proc. Computer Vision and Pattern Recognition (CVPR)}, pages
  2507--2516, 2019.

\bibitem{PASCAL_CONTEXT}
Roozbeh Mottaghi, Xianjie Chen, Xiaobai Liu, Nam-Gyu Cho, Sanja~Fidler
  Seong-Whan~Lee, Raquel Urtasun, and Alan Yuille.
\newblock The role of context for object detection and semantic segmentation in
  the wild.
\newblock In {\em Proc. Computer Vision and Pattern Recognition (CVPR)}, pages
  891--898, 2014.

\bibitem{MVD2017}
Gerhard Neuhold, Tobias Ollmann, Samuel Rota~Bul\`o, and Peter Kontschieder.
\newblock The mapillary vistas dataset for semantic understanding of street
  scenes.
\newblock In {\em Proc. International Conference on Computer Vision (ICCV)},
  2017.

\bibitem{scikit-learn}
F. Pedregosa, G. Varoquaux, A. Gramfort, V. Michel, B. Thirion, O. Grisel, M.
  Blondel, P. Prettenhofer, R. Weiss, V. Dubourg, J. Vanderplas, A. Passos, D.
  Cournapeau, M. Brucher, M. Perrot, and E. Duchesnay.
\newblock Scikit-learn: Machine learning in {P}ython.
\newblock {\em Journal of Machine Learning Research}, 12:2825--2830, 2011.

\bibitem{pohlen2017full}
Tobias Pohlen, Alexander Hermans, Markus Mathias, and Bastian Leibe.
\newblock Full-resolution residual networks for semantic segmentation in street
  scenes.
\newblock In {\em Proc. Computer Vision and Pattern Recognition (CVPR)}, pages
  4151--4160, 2017.

\bibitem{GTA5}
Stephan~R Richter, Vibhav Vineet, Stefan Roth, and Vladlen Koltun.
\newblock Playing for data: Ground truth from computer games.
\newblock In {\em Proc. European Conference on Computer Vision (ECCV)}, pages
  102--118, 2016.

\bibitem{U-Net}
Olaf Ronneberger, Philipp Fischer, and Thomas Brox.
\newblock U-net: Convolutional networks for biomedical image segmentation.
\newblock In {\em Proc. Medical Image Computing and Computer-Assisted
  Intervention (MICCAI)}, pages 234--241, 2015.

\bibitem{SYNTHIA}
German Ros, Laura Sellart, Joanna Materzynska, David Vazquez, and Antonio opez.
\newblock {The SYNTHIA Dataset}: A large collection of synthetic images for
  semantic egmentation of urban scenes.
\newblock In {\em Proc. Computer Vision and Pattern Recognition (CVPR)}, pages
  3234--3243, 2016.

\bibitem{saleh2018effective}
Fatemeh~Sadat Saleh, Mohammad~Sadegh Aliakbarian, Mathieu Salzmann, Lars
  Petersson, and Jose~M Alvarez.
\newblock Effective use of synthetic data for urban scene semantic
  segmentation.
\newblock In {\em Proc. European Conference on Computer Vision (ECCV)}, pages
  86--103. Springer, 2018.

\bibitem{sankaranarayanan2018learning}
Swami Sankaranarayanan, Yogesh Balaji, Arpit Jain, Ser Nam~Lim, and Rama
  Chellappa.
\newblock Learning from synthetic data: Addressing domain shift for semantic
  segmentation.
\newblock In {\em Proc. Computer Vision and Pattern Recognition (CVPR)}, pages
  3752--3761, 2018.

\bibitem{VGG}
Karen Simonyan and Andrew Zisserman.
\newblock Very deep convolutional networks for large-scale image recognition.
\newblock In {\em Proc. International Conference on Learning Representations
  (ICLR)}, 2015.

\bibitem{GAN_DA}
Yi-Hsuan Tsai, Wei-Chih Hung, Samuel Schulter, Kihyuk Sohn, Ming-Hsuan Yang,
  and Manmohan Chandraker.
\newblock Learning to adapt structured output space for semantic segmentation.
\newblock In {\em Proc. Computer Vision and Pattern Recognition (CVPR)}, pages
  7472--7481, 2018.

\bibitem{BDD100K}
Fisher Yu, Haofeng Chen, Xin Wang, Wenqi Xian, Yingying Chen, Fangchen Liu,
  Vashisht Madhavan, and Trevor Darrell.
\newblock Bdd100k: A diverse driving dataset for heterogeneous multitask
  learning.
\newblock In {\em Proc. Computer Vision and Pattern Recognition (CVPR)}, pages
  2636--2645, 2020.

\bibitem{zendel2018wilddash}
Oliver Zendel, Katrin Honauer, Markus Murschitz, Daniel Steininger, and Gustavo
  Fernandez~Dominguez.
\newblock Wilddash-creating hazard-aware benchmarks.
\newblock In {\em Proc. European Conference on Computer Vision (ECCV)}, pages
  402--416, 2018.

\bibitem{zhang2017curriculum}
Yang Zhang, Philip David, and Boqing Gong.
\newblock Curriculum domain adaptation for semantic segmentation of urban
  scenes.
\newblock In {\em Proc. International Conference on Computer Vision (ICCV)},
  pages 2020--2030, 2017.

\bibitem{PSP}
Hengshuang Zhao, Jianping Shi, Xiaojuan Qi, Xiaogang Wang, and Jiaya Jia.
\newblock Pyramid scene parsing network.
\newblock In {\em Proc. Computer Vision and Pattern Recognition (CVPR)}, pages
  2881--2890, 2017.

\bibitem{zou2019confidence}
Yang Zou, Zhiding Yu, Xiaofeng Liu, BVK Kumar, and Jinsong Wang.
\newblock Confidence regularized self-training.
\newblock In {\em Proc. International Conference on Computer Vision (ICCV)},
  pages 5982--5991, 2019.

\bibitem{zou2018unsupervised}
Yang Zou, Zhiding Yu, BVK Vijaya~Kumar, and Jinsong Wang.
\newblock Unsupervised domain adaptation for semantic segmentation via
  class-balanced self-training.
\newblock In {\em Proc. European Conference on Computer Vision (ECCV)}, pages
  289--305, 2018.

\end{thebibliography}
}

\end{document}